\documentclass[preprint,12pt, a4paper]{elsarticle}

\usepackage{amssymb}

\usepackage{lineno}
\usepackage{boldline}
\usepackage[caption=false]{subfig}
\usepackage{float}
\usepackage{hyperref}
\usepackage{booktabs} 
\usepackage{algorithm}%
\usepackage{algpseudocode}%
\usepackage{amsmath}

\restylefloat{table}

\journal{SoftwareX}

\begin{document}

\begin{frontmatter}



\title{GTOPX Space Mission Benchmarks}


\author[label1]{Martin Schlueter*}%
\author[label2]{ Mehdi Neshat}
\author[label3]{Mohamed Wahib}
\author[label4]{Masaharu Munetomo}
\author[label5]{Markus Wagner}

\address[label1]{Information Initiative Center, Hokkaido University, Sapporo 060-811, Japan, \texttt{schlueter@midaco-solver.com}}

\address[label2]{Optimisation and Logistics Group, School of Computer Science, The University of Adelaide, Australia, \texttt{mehdi.neshat@adelaide.edu.au}}

\address[label3]{AIST-Tokyo Tech Real World Big-Data Computation Open Innovation Laboratory, National Institute of Advanced Industrial Science and Technology Tokyo, Japan, \texttt{mohamed.attia@aist.go.jp}}

\address[label4]{Information Initiative Center,	Hokkaido University, Sapporo 060-811, Japan,
   \texttt{munetomo@iic.hokudai.ac.jp}}

\address[label5]{Optimisation and Logistics Group,
  School of Computer Science,
  The University of Adelaide,
   Australia,
  \texttt{markus.wagner@adelaide.edu.au} }
  
 
\begin{abstract}
This contribution introduces the GTOPX space mission benchmark collection, which is an extension of the GTOP database published by the European Space Agency (ESA). The term GTOPX stands for \underline{G}lobal \underline{T}rajectory \underline{O}ptimization \underline{P}roblems with e\underline{X}tension. GTOPX consists of ten individual benchmark instances representing real-world interplanetary space trajectory design problems. In regard to the original GTOP collection, GTOPX includes three new problem instances featuring mixed-integer and multi-objective properties. GTOPX enables a simplified user-handling, unified benchmark function call and some minor bug corrections to the original GTOP implementation. Furthermore, GTOPX is linked from original C++ source code to Python and Matlab based on dynamic link libraries, assuring computationally fast and accurate reproduction of the benchmark results in all programming languages. We performed a comprehensive landscape analysis to characterize the properties of the fitness landscape of GTOPX benchmarks. Space mission trajectory design problems as those represented in GTOPX are known to be highly non-linear and difficult to solve. The GTOPX collection therefore aims in particular at researchers wishing to put advanced (meta)heuristic and hybrid optimization algorithms to the test.

\end{abstract}
\sloppy

\end{frontmatter}










\section{Motivation and significance}
\label{sec:intro} 
\noindent
The optimal design of interplanetary space mission trajectories is an active and challenging research
area in aerospace and related communities, such as the evolutionary and metaheuristic optimization community. Since 2005, the European Space Agency (ESA) maintained a set of real-world space mission trajectory design problems in form of numerical black-box optimization problems, known as the GTOP database~\cite{gtop}. Note that the term ``black-box'' refers here to an optimization problem, where the actual problem formulation is unknown, inaccessible or unrelated to the optimizer. These kind of black-box problems frequently occur in many areas, where complex computer simulations are applied. Given the usefulness and success of the GTOP endeavour in the past, here we present an extended and refurbished version, named GTOPX, as continuation to the no longer maintained GTOP database. 

While all problems of the original GTOP database were single-objective and continuous in the nature of the search space domain, the GTOPX collection includes three new problem instances featuring mixed-integer and multi-objective problem properties. The general mathematical form of the considered optimization problem in GTOPX is given as multi-objective Mixed-integer Nonlinear Programming (MINLP):

\begin{equation}
\label{minlp}
\begin{split}
\text{Minimize }~~~f_{i}(x,y)~~~~~~~~~~~~&(x\in\mathbb{R}^{n_{con}},~y\in\mathbb{N}^{n_{int}},~n_{con},~n_{int}\in\mathbb{N})\\
~\\
\text{subject to:}~~
g_{i}(x,y)~\geq~0,~~~&i~=~1,...,m\in\mathbb{N}\\
x_{l} \leq~ x ~\leq x_{u}~~~&(x_{l},~x_{u}\in\mathbb{R}^{n_{con}})\\
y_{l} \leq~ y ~\leq y_{u}~~~&(y_{l},~y_{u}\in\mathbb{N}^{n_{int}})
\end{split}
\end{equation}

where $f_{i}(x,y)$ and $g_{i}(x,y)$ denote the objective and constraint functions depending on continuous ($x$) and discrete ($y$) decision variables, which are box-constrained to some lower ($x_{l}$, $y_{l}$) and upper  ($x_{u}$, $y_{u}$) bounds. The ten individual benchmark instances of GTOPX are listed in Table \ref{tab:GTOPX} with their name, the number of objectives, variables and constraints together with the best known objective function value $f(x,y)$. Note that due to the many years these benchmarks have been available and been tested, it is believed that all listed solutions are essentially converged.

\begin{table}[!h] 
\centering
\caption{GTOPX Benchmark Instances (see \cite{gtop} for solution references)}
\label{tab:GTOPX}
\scalebox{0.85}{
 \begin{tabular}{|clcccc|}
 \hline 
 No.& Name & Objectives & Variables & Constraints & Best known $f(x,y)$  \\
  \hline 
  1 & Cassini1            & 1 & 6  & 4 & 4.9307 \\
  2 & Cassini2            & 1 & 22 & 0 & 8.3830 \\
  3 & Messenger (reduced) & 1 & 18 & 0 & 8.6299 \\
  4 & Messenger (full)    & 1 & 26 & 0 & 1.9579 \\
  5 & GTOC1               & 1 & 8  & 6 & -1581950.0 \\
  6 & Rosetta             & 1 & 22 & 0 &  1.3434 \\
  7 & Sagas               & 1 & 12 & 2 &  18.1877 \\
  8 & Cassini1-MINLP      & 1 & 10 & 4 & 3.5007 \\
  9 & Cassini1-MO         & 2 & 6  & 5 & na \\
  10 & Cassini1-MO-MINLP  & 2 & 10 & 5 & na \\                    
 \hline 
 \end{tabular} 
 }
\end{table} 

The GTOP database has attracted a significant amount of attention and results have been published, see for example \cite{Addis,Ampatzis,biazzini,Biscani,Gad,Gregoire,Gruber,Henderson,Islam,Izzo,Lancinskas,Musegaas,Stracquadanio,vinko}. Note that the majority of publications discussing results on GTOP focus on only one or a few problem instances. This is due to the difficulty of these problems, which often require many millions of function evaluations to allow an optimization algorithm to converge to the best known solution. This difficulty is what makes this collection of benchmark problems interesting and a real challenge, even requiring the use of massively distributed computing power by supercomputers in the most difficult cases (see Shuka~\cite{Shuka} or Schlueter~\cite{schl2017}).

While the focus of the original GTOP database was rather on the nature of the application itself, the GTOPX collection aims to address the broad community of numerical optimization researchers. It does so by a simplified and more user-friendly source code base, that allows easy execution of the benchmarks in three programming languages: C/C++, Python and Matlab. The entire source code base of GTOPX has been compressed into a single file (namely ``gtopx.cpp'') that is linked via dynamic link libraries (DLL) into the Python and Matlab language. The linking of a pre-compiled DLL has the significant advantage of computation speed and accurate reproduction of results among all considered programming languages, in contrast to a native re-implementation in other languages, which is computationally slower and error prone.

The function call to the individual GTOPX benchmark instances has been generalized, so that any problem instance can easily be accessed with only switching one integer parameter (namely the benchmark number). The generalized call to the GTOPX problem functions in each language is given as follows:

\begin{table}[H]
\centering
\begin{tabular}{lcl}
C/C++   &~~~:~~~& gtopx( benchmark, f, g, x ); \\
Python  &~~~:~~~& [ f, g ] = gtopx( benchmark, x, o,n,m ) \\
Matlab  &~~~:~~~& [ f, g ] = gtopxmex( benchmark, x ) \\
\end{tabular}
\end{table}

In above table, \textit{f} denotes the objective function value(s), \textit{g} denotes the constraint function values, \textit{x} denotes the vector of decision variables and \textit{o, m} and \textit{n} denote their corresponding dimensions. All GTOPX benchmark instances are thread-safe for execution. This means that the GTOPX problems can be executed in parallel, which is a highly desired feature in modern optimization algorithm design. The GTOPX source code files are free software and published under the GPL license~\cite{GPL}.

The remainder of this paper is structured as follows. Section~\ref{sec:2} describes each of the ten individual instances from Table~\ref{tab:GTOPX} in detail. Section~\ref{sec:Landscape_Analysis} contains a comprehensive fitness landscape analysis of the GTOPX single objective benchmarks to provide a better perception of a proper optimisation algorithm selection. A brief summary is given with some conclusive statements. This paper is to provide researchers with a manual and reference to the GTOPX benchmark software.
\section{Software description}
\label{sec:2}


This section gives detailed information to the ten GTOPX benchmark instances. 
Note that the definitions of the first seven instances\footnote[1]{Note that an individual benchmark problem is also called an \textit{instance} in the optimization community.} are taken from Schlueter~\cite{schlueter:2014}, while the last three instances are new.



\subsection{ Cassini1}

The Cassini1 benchmark models an interplanetary space mission to Saturn.
The objective of the mission is to get captured by Saturn's gravity into an orbit having a pericenter radius of 108,950 km and an eccentricity of 0.98. The sequence of fly-by planets for this mission is given by Earth-Venus-Venus-Earth-Jupiter-Saturn, whereas the first item is the start planet and the last item is the final target.
The objective function of this benchmark is to minimize the total velocity change $\Delta V$ accumulated during the mission, including the launch and capture manoeuvre. 
The benchmark involves six decision variables:

\begin{table}[!h] 
\centering
\caption{Description of optimization variables for Cassini1}
\label{var:cassini1}
 \begin{tabular}{cc}
 \hlineB{3} 
 Variable & Description  \\
  \hline 
  1 & Initial day measured from 1-Jan 2000 \\
  2 $\sim$ 6 & Time interval between events (e.g. departure, fly-by, capture) \\
 \hline 
 \end{tabular} 
\end{table}

This benchmark further considers four constraints, which impose an upper limit on the pericenters for the four fly-by maneuvers. The best known solution to this benchmark has an objective function value of $f(x) = 4.9307$, and the respective vector of solution decision variables $x$ is available online \cite{GTOPX}.

\subsection{ Cassini2}

The Cassini2 benchmark models an interplanetary space mission to Saturn, including deep space maneuvers (DSM) and is therefore considerably more difficult than its counterpart benchmark Cassini1. The sequence of fly-by planets for this mission is given by Earth-Venus-Venus-Earth-Jupiter-Saturn, where the first item is the start planet and the last item is the final target. The objective of this benchmark is again to minimize the total $\Delta V$ accumulated during the full mission; however, the aim here is a rendezvous, while in Cassini1 the aim is an orbit insertion. The benchmark involves 22 decision variables: 

\begin{table}[!h] 
\centering
\caption{Description of optimization variables for Cassini2}
\label{var:cassini2}
 \begin{tabular}{cc}
 \hlineB{3} 
 Variable & Description  \\
  \hline 
  1 & Initial day measured from 1-Jan 2000 \\
  2 & Initial excess hyperbolic speed (km/Sec) \\  
  3 & Angles of the hyperbolic excess velocity (polar coordinate frame) \\
  4 & Angles of the hyperbolic excess velocity (polar coordinate frame) \\  
  5 $\sim$ 9 & Time interval between events (e.g. departure, fly-by, capture) \\
 10 $\sim$ 14 & Fraction of the time interval after which DSM occurs \\
  15 $\sim$ 18 & Radius of flyby (in planet radii) \\
  19 $\sim$ 22 & Angle measured in B plane of the planet approach vector\\
 \hline 
 \end{tabular} 
\end{table} 

The best known solution to this benchmark has an objective function value of $f(x) = 8.3830$, and the respective vector of solution decision variables $x$ is available online~\cite{GTOPX}.

\subsection{Messenger (reduced)}  

This benchmark models an interplanetary space mission to Mercury and does not include resonant fly-bys at Mercury. The sequence of fly-by planets for this mission is given by Earth-Earth-Venus-Venus-Mercury, where the first item is the start planet and the last item is the final target. The objective of this benchmark to be minimized is again the total $\Delta V$ accumulated during the mission. The benchmark involves 18 decision variables:

\begin{table}[!h] 
\centering
\caption{Description of optimization variables for Messenger (reduced)}
\label{var:messenger}
 \begin{tabular}{cc}
 \hlineB{3} 
 Variable & Description  \\
  \hline 
  1 & Initial day measured from 1-Jan 2000 \\
  2 & Initial excess hyperbolic speed (km/sec) \\  
  3 & Angles of the hyperbolic excess velocity (polar coordinate frame) \\
  4 & Angles of the hyperbolic excess velocity (polar coordinate frame) \\  
  5 $\sim$ 8 & Time interval between events (e.g. departure, fly-by, capture) \\
  9 $\sim$ 12 & Fraction of the time interval after which DSM occurs \\
  13 $\sim$ 15 & Radius of fly-by (in planet radii) \\
  16 $\sim$ 18 & Angle measured in B plane of the planet approach vector\\
 \hline 
 \end{tabular} 
\end{table} 
The best known solution to this benchmark has an objective function value of $f(x) = 8.6299$, and the respective vector of solution decision variables $x$ is available online~\cite{GTOPX}.

\subsection{Messenger (full)}

This benchmark models an interplanetary space mission to Mercury, including resonant fly-bys at Mercury. The sequence of fly-by planets for this mission is given by Earth-Venus-Venus-Mercury-Mercury-Mercury-Mercury. The objective of this benchmark to be minimized is -- as before -- the total $\Delta V$ accumulated during the full mission. The benchmark involves 26 decision variables: 

\begin{table}[!h] 
\centering
\caption{Description of optimization variables for Messenger (full)}
\label{var:messfull}
 \begin{tabular}{cc}
 \hlineB{3} 
 Variable & Description  \\
  \hline 
  1 & Initial day measured from 1-Jan 2000 \\
  2 & Initial excess hyperbolic speed (km/sec) \\  
  3 & Angles of the hyperbolic excess velocity (polar coordinate frame) \\
  4 & Angles of the hyperbolic excess velocity (polar coordinate frame) \\  
  5 $\sim$ 10 & Time interval between events (e.g. departure, fly-by, capture) \\
  11 $\sim$ 16 & Fraction of the time interval after which DSM occurs \\
  17 $\sim$ 21 & Radius of fly-by (in planet radii) \\
  22 $\sim$ 26 & Angle measured in B plane of the planet approach vector\\
 \hline 
 \end{tabular} 
\end{table}

The best known solution to this benchmark has an objective function value of $f(x) = 1.9579$, and the vector of solution decision variables $x$ is available online~\cite{GTOPX}.

\clearpage

\subsection{GTOC1}

This benchmark models a multi gravity assist space mission to asteroid TW229. The mission model drew inspiration from the first edition of the Global Trajectory Optimisation Competition (GTOC) held by ESA in 2007~\cite{izzo2007}. The objective of the mission is to maximize the change in the semi-major axis of the asteroid orbit. The sequence of fly-by planets for this mission is given by Earth-Venus-Earth-Venus-Earth-Jupiter-Saturn-TW229. This benchmark involves eight decision variables:

\begin{table}[!h] 
\centering
\caption{Description of optimization variables for GTOC1}
\label{var:GTOC1}
 \begin{tabular}{cc}
 \hlineB{3} 
 Variable & Description  \\
  \hline 
  1 & Initial day measured from 1-Jan 2000 \\ 
  2 $\sim$ 8 & Time interval between events (e.g. departure, fly-by, capture) \\
 \hline 
 \end{tabular} 
\end{table}

This benchmark further considers four constraints, which impose an upper limit on the pericenters for the four fly-by maneuvers. The best known solution to this benchmark has an objective function value of $f(x) = -1581950.0$, and the vector of solution decision variables $x$ is available online~\cite{GTOPX}.

\subsection{Rosetta}

The Rosetta benchmark models multi gravity assist space mission to comet 67P/Churyumov-Gerasimenko, including deep space maneuvers (DSM). 
The sequence of fly-by planets for this mission is given by Earth-Earth-Mars-Earth-Earth-67P. The objective of this benchmark is to minimize the total $\Delta V$ accumulated during the mission. The benchmark involves 22 decision variables:

\begin{table}[!h] 
\centering
\caption{Description of optimization variables for Rosetta}
\label{var:rosetta}
 \begin{tabular}{cc}
 \hlineB{3} 
 Variable & Description  \\
  \hline 
  1 & Initial day measured from 1-Jan 2000 \\
  2 & Initial excess hyperbolic speed (km/sec) \\  
  3 & Angles of the hyperbolic excess velocity (polar coordinate frame) \\
  4 & Angles of the hyperbolic excess velocity (polar coordinate frame) \\  
  5 $\sim$ 9 & Time interval between events (e.g. departure, fly-by, capture) \\
 10 $\sim$ 14 & Fraction of the time interval after which DSM occurs \\
  15 $\sim$ 18 & Radius of fly-by (in planet radii) \\
  19 $\sim$ 22 & Angle measured in B plane of the planet approach vector\\
 \hline 
 \end{tabular} 
\end{table}

The best known solution to this benchmark has an objective function value of $f(x) = 1.3434$, and the vector of solution decision variables $x$ is available online~\cite{GTOPX}.
\subsection{Sagas}

This benchmark is described as a $\Delta V$-EGA manoeuvre to fly-by Jupiter and reach 50AU. The sequence of fly-by planets for this mission is given by Earth-Earth-Jupiter, where the first item is the start planet and the last item is the final planet. The objective of this benchmark is to minimize the total $\Delta V$ accumulated during the mission. The benchmark involves 12 decision variables described in Table~\ref{var:sagas}.

\begin{table}[!h] 
\centering
\caption{Description of optimization variables for Sagas}
\label{var:sagas}
 \begin{tabular}{cc}
 \hlineB{3} 
 Variable & Description  \\
  \hline 
  1 & Initial day measured from 1-Jan 2000 \\
  2 & Initial excess hyperbolic speed (km/sec) \\  
  3 & Angles of the hyperbolic excess velocity (polar coordinate frame) \\
  4 & Angles of the hyperbolic excess velocity (polar coordinate frame) \\  
  5 $\sim$ 6 & Time interval between events (e.g. departure, fly-by, capture) \\
  7 $\sim$ 8 & Fraction of the time interval after which DSM occurs \\
  9 $\sim$ 10 & Radius of fly-by (in planet radii) \\
  11 $\sim$ 12 & Angle measured in B plane of the planet approach vector\\
 \hline 
 \end{tabular} 
\end{table}

This benchmark further considers two constraints, which impose an upper limit on the on-board fuel and launcher performance. The best known solution to this benchmark has an objective function value of $f(x) = 18.1877$, and the respective vector of solution decision variables $x$ is available online \cite{GTOPX}.

\subsection{Cassini1-MINLP}

This new benchmark is a mixed-integer extension of the Cassini1 instance. While in the original Cassini1 instance the sequence of fly-by planets is fixed as Venus-Venus-Earth-Jupiter, the Cassini1-MINLP considers all four fly-by planets as discrete decision variable. Each planet of the solar system (plus the dwarf planet Pluto) is a feasible choice for any of the four fly-by planets. Table~\ref{planets} lists all possible choices for fly-by planets together with their numerical value, to be used as variable for the GTOPX solution vector $x$. Table~\ref{var:cassini1minlp} lists the ten decision variables.

\begin{table}[h!]
\centering \caption{Possible choices of fly-by planets}
\label{planets}
\begin{tabular}{cc}
\hline
~~~Value~~~  &         ~~~~~~Planet~~~~~~ \\
\hline
1 & Mercury \\
2 & Venus \\
3 & Earth \\
4 & Mars \\
5 & Jupiter \\
6 & Saturn \\
7 & Uranus \\
8 & Neptune \\
9 & Pluto \\
    \hline
\end{tabular}
\end{table}

\begin{table}[!h] 
\centering
\caption{Description of optimization variables for Cassini1-MINLP}
\label{var:cassini1minlp}
 \begin{tabular}{cc}
 \hlineB{3} 
 Variable & Description  \\
  \hline 
  1 & Initial day measured from 1-Jan 2000 \\
  2 $\sim$ 6 & Time interval between events (e.g. departure, fly-by, capture) \\
  7 $\sim$10 & fly-by planet (discrete value, see Table \ref{planets}) \\  
 \hline 
 \end{tabular} 
\end{table} 
The Cassini1-MINLP problem has been previously investigated with numerical results in Schlueter~\cite{schl2019}. In that paper it was revealed, that this benchmark has a strong local minimum corresponding to the fly-by planet sequence \{Earth, Earth, Earth, Jupiter\}. This local minimum appeared to be the second best solution (in regard to the combinatorial part) and holds an objective function value of $f(x) = 3.6307$. The best known solution to Cassini1-MINLP has the fly-by planet sequence \{Earth, Venus, Earth, Jupiter\} and corresponds to an objective function value of $f(x) = 3.5007$.

The decisive nature of the above described local minima makes the Cassini1-MINLP exceptionally hard to solve. It is therefore noteworthy that the Cassini1 instance and the Cassini1-MINLP instance are significantly different in their complexity and difficulty.

Note that the technical modification in the source code of the GTOP database, necessary to enable the integer choice of \textit{fly-by} planets, is given in the C++ function \texttt{cassini1minlp}. In such function, the following original source code line:

{\small
\texttt{sequence\_[CASSINI\_DIM] = \{3,2,2,3,5,6\}}
}

\noindent
is replaced with the following:

{\small
\texttt{sequence\_[CASSINI\_DIM] = \{3,{$x_{7}$},{$x_{8}$},{$x_{9}$},{$x_{10}$},6\}}
}

\noindent
where the sequence elements {$x_{7}$},{$x_{8}$},{$x_{9}$},{$x_{10}$} represent the integer variables. Note that the first and last entry in the above source code sequence represent the start and target planet, which is in this case Earth ($\rightarrow$ start) and Saturn ($\rightarrow$ target). Therefore the first and last entry in the original sequence remain unaffected by the above modification.

\clearpage

\subsection{ Cassini1-MO}

This benchmark is a multi-objective extension of the Cassini1 instance. While in the original Cassini1 instance only one objective (the total $\Delta V$) was considered, the Cassini1-MO benchmark considers as the second objective the total time of flight (measured in days). The decision variables for Cassini1-MO are identical to the Cassini1 instance:


\begin{figure}[!h]
   \centering
   \caption{Pareto front of Cassini1-MO}
   \includegraphics[width=0.9\textwidth]{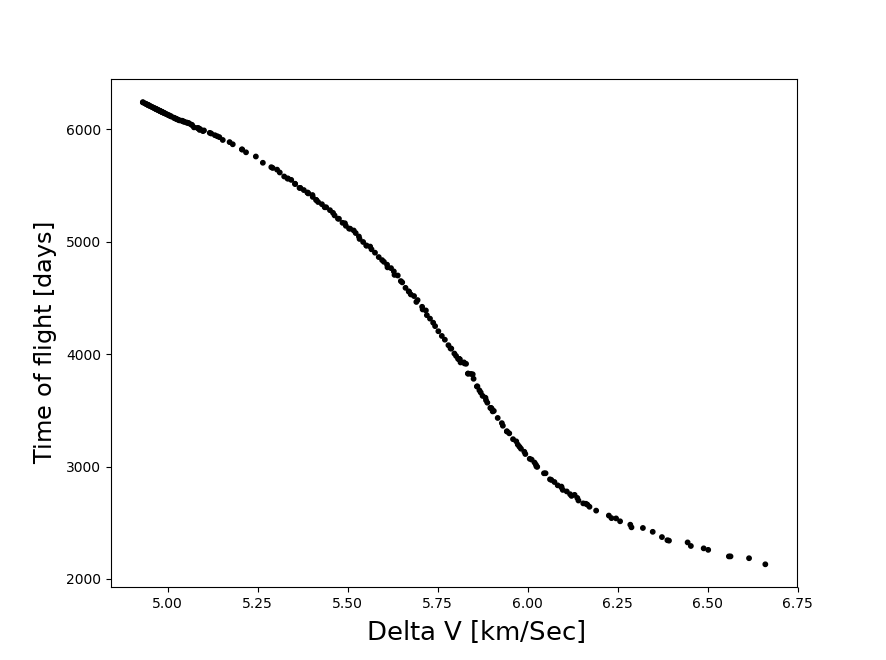}
   \label{fig:mo}
\end{figure}
\begin{table}[!h] 
\centering
\caption{Description of optimization variables for Cassini1-MO}
\label{var:cassini1}
 \begin{tabular}{cc}
 \hlineB{3} 
 Variable & Description  \\
  \hline 
  1 & Initial day measured from 1-Jan 2000 \\
  2 $\sim$ 6 & Time interval between events (e.g. departure, fly-by, capture) \\
 \hline 
 \end{tabular} 
\end{table}

The Cassini1-MO has one more constraint than Cassini1. This additional constraint bounds the objective space to solutions with a first objective-function (Delta V [m/s]) value of $f_{1}(x) \leq 7.0$. This additional constraint is introduced to focus the area of the Pareto front to the truly challenging region of low fuel-consuming space mission trajectories. 

There is yet no comprehensive analysis conducted on this benchmark, but an approximation of the Pareto front was obtained by the MIDACO solver~\cite{schlueter:2013a} and is displayed in Figure~\ref{fig:mo}. There, the Pareto front of Cassini1-MO reveals a very distinctive non-separated shape, exhibiting convex (right half of Figure~\ref{fig:mo}) as well as non-convex (left half of Figure~\ref{fig:mo}) areas. The set of non-dominated solutions is available online at~\cite{GTOPX} as a text file.

Note that while the benchmark definition allows feasible points with $f_{1}(x) \leq 7.0$, there appears to exist only feasible non-dominated solutions with a value of around $f_{1}(x) \leq 6.7$. 

Note that in contrast to single-optimization, there does not exist a commonly agreed on performance measure in multi-objective optimization. Among the most widely used methods to measure the performance in multi-objective optimization are the hyper-volume (HV) indicator and the general and inverse general distance (GD $\&$ IGD) indicators~\cite{chand2015emo}. Thus the authors provide only the raw data of the Pareto front approximation and interested users must choose their own choice of multi-objective performance measure in order to compare specific algorithms.

\subsection{ Cassini1-MO-MINLP}

The Cassini1-MO-MINLP benchmark combines the previous two extensions of Cassini1-MINLP and Cassini1-MO. This benchmark is therefore a multi-objective mixed-integer problem and very challenging to solve.
The decision variables are the same as for Cassini1-MINLP, and the fly-by planet choices are listed in Table \ref{planets}:


\begin{table}[!h] 
\centering
\caption{Description of optimization variables for Cassini1}
 \begin{tabular}{cc}
 \hlineB{3} 
 Variable & Description  \\
  \hline 
  1 & Initial day measured from 1-Jan 2000 \\
  2 $\sim$ 6 & Time interval between events (e.g. departure, fly-by, capture) \\
  7 $\sim$10 & Fly-By planet (discrete value, see Table \ref{planets}) \\  
 \hline 
 \end{tabular} 
\end{table} 

As in Cassini1-MO, the first objective of Cassini1-MO-MINLP is the total $\Delta V$ and the second objective is the total time of flight of the space mission. Like the Cassini1-MO benchmark the Cassini1-MO-MINLP has an additional constraint on the objective space. However, in contrast to Cassini1-MO this benchmark bounds the objective space to solutions with a first objective-function (Delta V [m/s]) value of $f_{1}(x) \leq 6.0$ instead of a value of 7.0, as the changed search space landscape allows for better solutions than the Cassini1-MO benchmark.

\begin{figure}[!h]
   \centering
   \caption{Pareto front of Cassini1-MO-MINLP}    
   \includegraphics[width=0.8\textwidth]{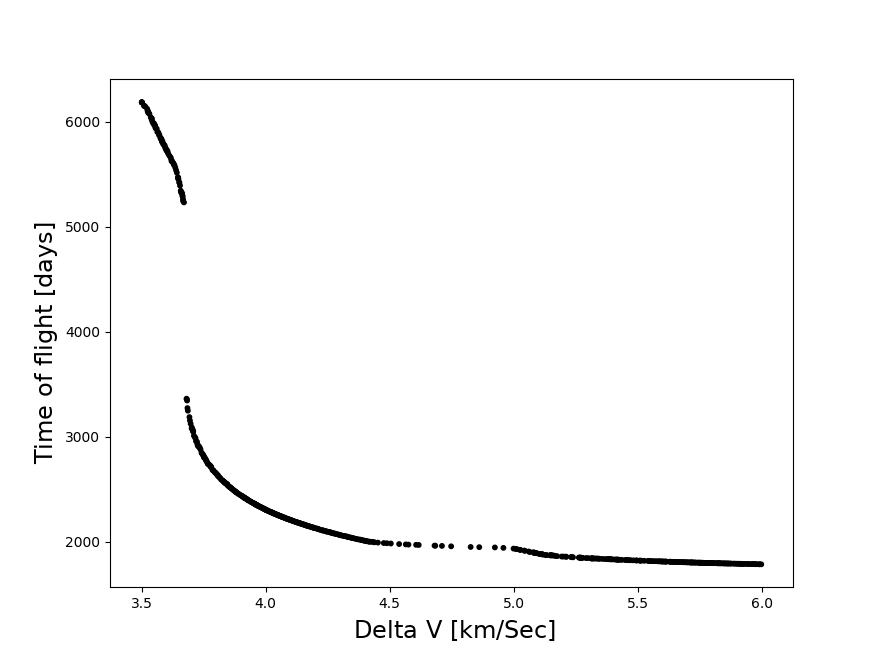}
   \label{fig:mo-minlp}
\end{figure}

The Pareto front (as approximated by the MIDACO solver in \cite{schlueter:2013a}, see Figure~\ref{fig:mo-minlp}) of Cassini1-MO reveals a distinctive separated shape, where it is to note that the upper left part is mostly corresponding to solutions with the best\footnote[1]{The best known integer sequence refers here to the value of the corresponding first objective function value.} known integer sequence while the lower part is corresponding to the second best known integer sequence. 
The full set of non-dominated solutions is also available online at \cite{GTOPX} as a text file. 

\section{Landscape Analysis}
\label{sec:Landscape_Analysis}

Landscape analysis methods can be employed to characterize the features of fitness landscapes by representing the levels of ruggedness, smoothness, multi-modality and neutrality~\cite{munoz2014exploratory}. 
Characteristics like these can be exploited by algorithm designers in the construction of an algorithm, but also in an online fashion by algorithms themselves. For example, when fitness values vary very little in a particular region (and the algorithm might be stuck on a plateau), an optimization method can adjust its parameters (like mutation step size) to generate candidates that are further away from the current solutions. 

In the following, we use two ways of characterising the benchmarks. First, we sample uniformly-at-random around the best-known solutions in a way that is similar to those used by some local search algorithms, and we do so with the goal of simulating how a local search would perceive the area around the best solutions -- we use parallel coordinate plots for the visualisation due to the high dimensionality. Second, we use systematic grid searches for a holistic picture of the entire problem -- we use 3D plots for the visualisation.


First, we investigate the local sampling around the best trajectories known. 
We sample the neighbourhood in a way similar to those employed by some local search approaches: (1) the mutation probability of each individual value is $\frac{1}{N}$, where $N$ is the number of decision variables, (2) we sample from a normal distribution around the best-known solution with $\sigma_i=\frac{Ub_i-Lb_i}{3}$, where $Ub$ and $Lb$ are the upper and lower bounds of the $i^{th}$ variable, and (3) we do so one million times. 

\begin{figure}[!h]
\centering
 \subfloat[]{
 \includegraphics[clip,height=3cm,width=1.0\columnwidth]{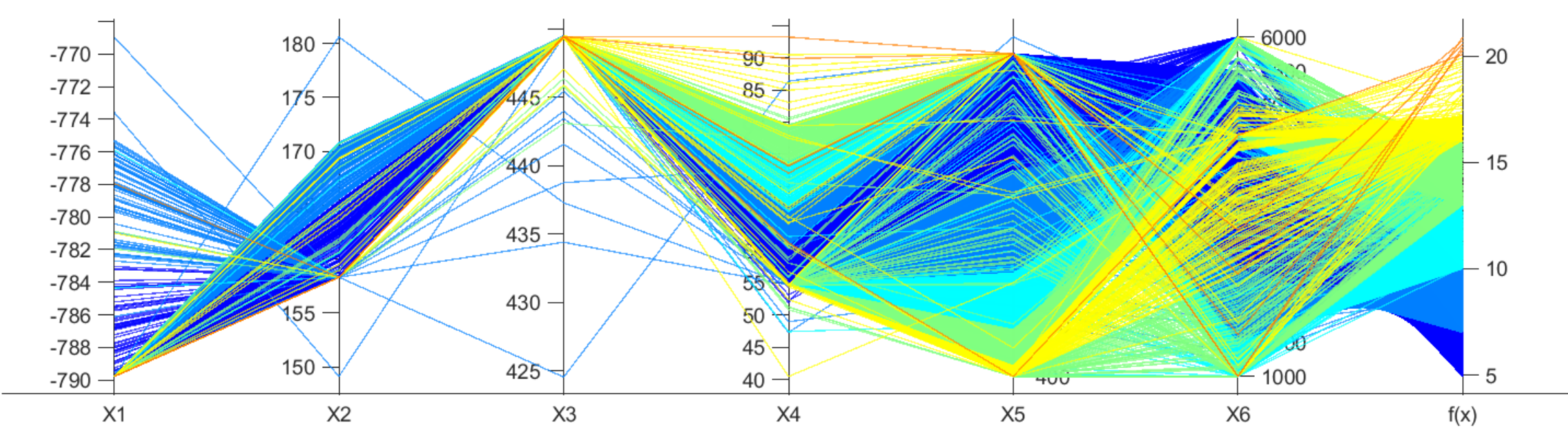}}\\
  \subfloat[]{
 \includegraphics[clip,height=3cm,width=0.96\columnwidth]{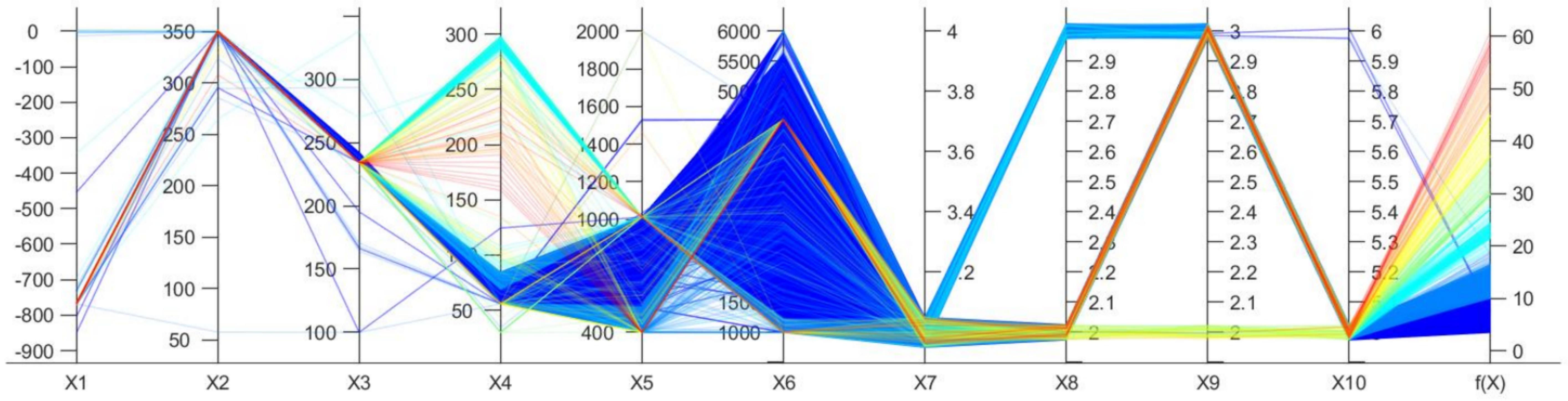}}\\
 \subfloat[]{
 \includegraphics[clip,height=3cm,width=1.0\columnwidth]{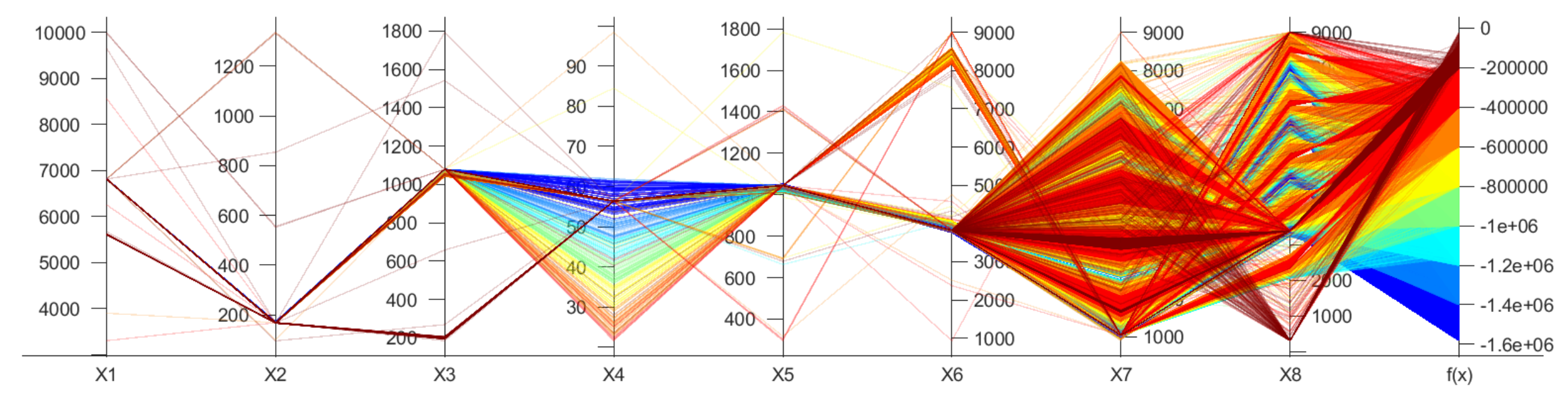}}\\ 

\caption{Parallel coordinate plots of (a) Cassini1, (b) Cassini1-MINLP and (c) GTOC1. The color of each line corresponds to the objective function value $f(x)$ given at the right side of each plot. Because all GTOPX benchmarks are to be minimized, a cold color (e.g. blue) indicates a better good solution, while a warm color (e.g. red) indicates a bad solution.}%
 \label{fig:parallel_cooperate_1}%
\end{figure}

\begin{figure}[!h]
\centering
 \subfloat[]{
 \includegraphics[clip,height=2.7cm,width=1.0\columnwidth]{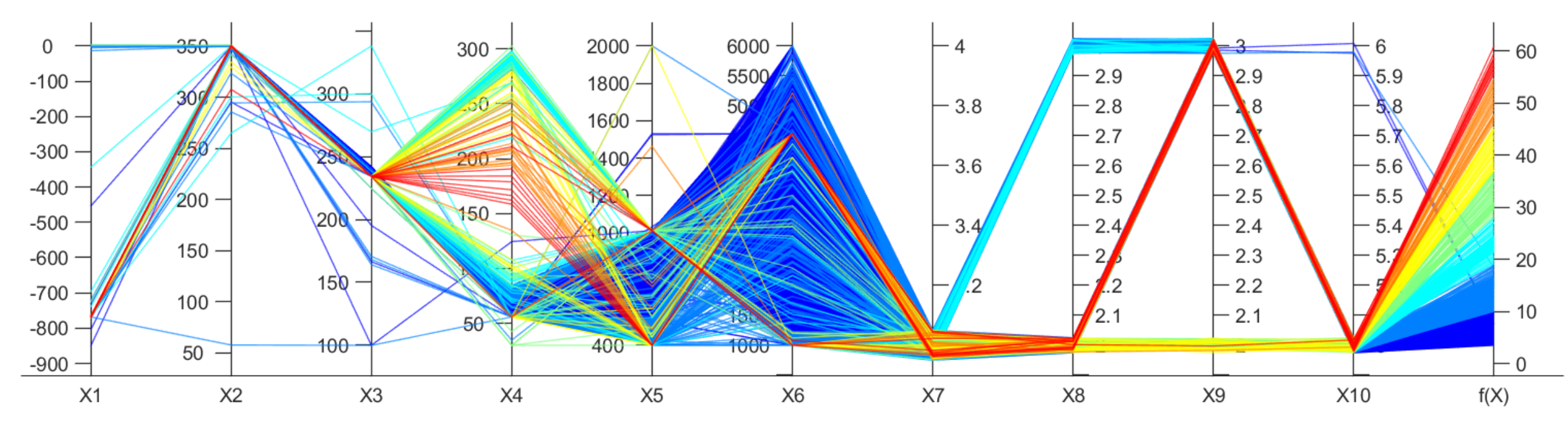}}\\
 \subfloat[]{
 \includegraphics[clip,height=2.7cm,width=1.0\columnwidth]{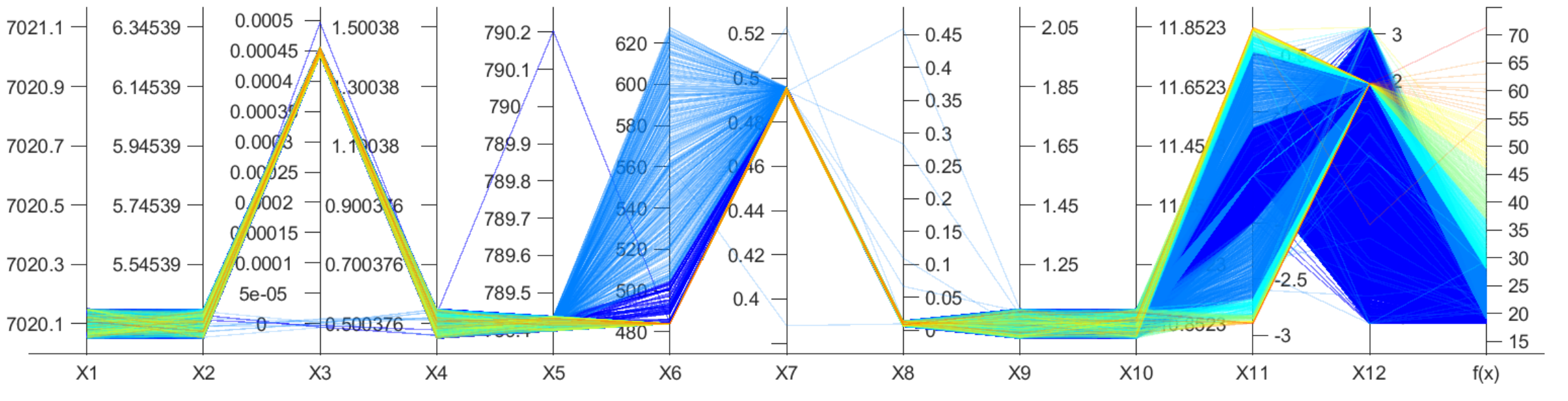}}\\
 \subfloat[]{
 \includegraphics[clip,height=2.7cm,width=1.0\columnwidth]{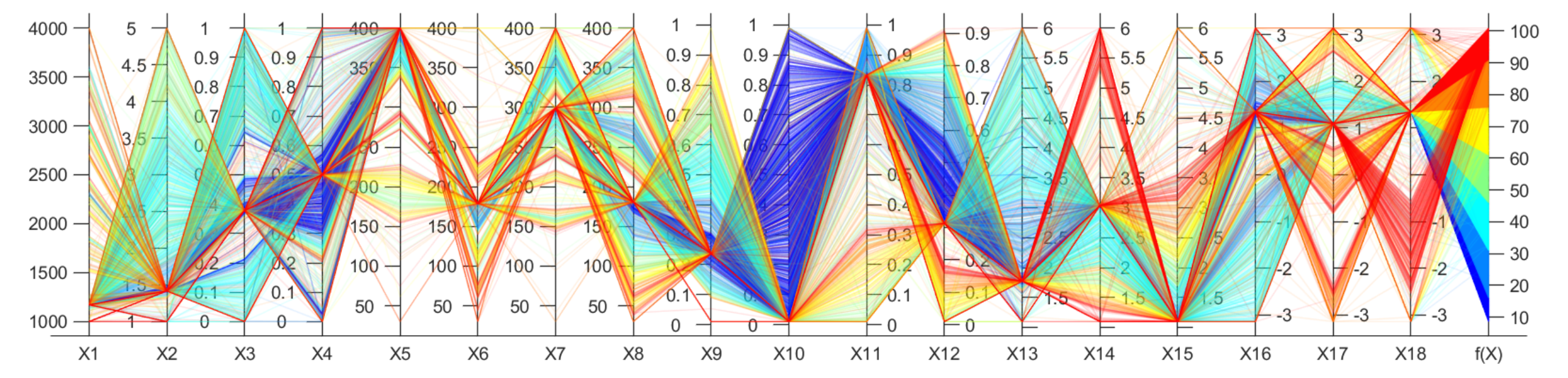}}\\ 
 \subfloat[]{
 \includegraphics[clip,height=2.7cm,width=1.0\columnwidth]{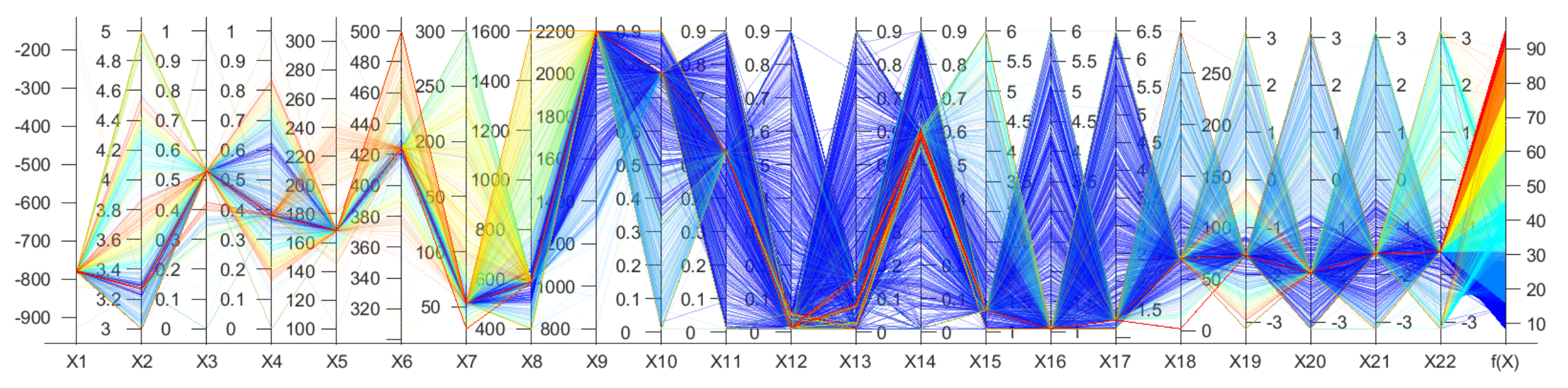}}\\
\subfloat[]{
\includegraphics[clip,height=2.7cm,width=1.0\columnwidth]{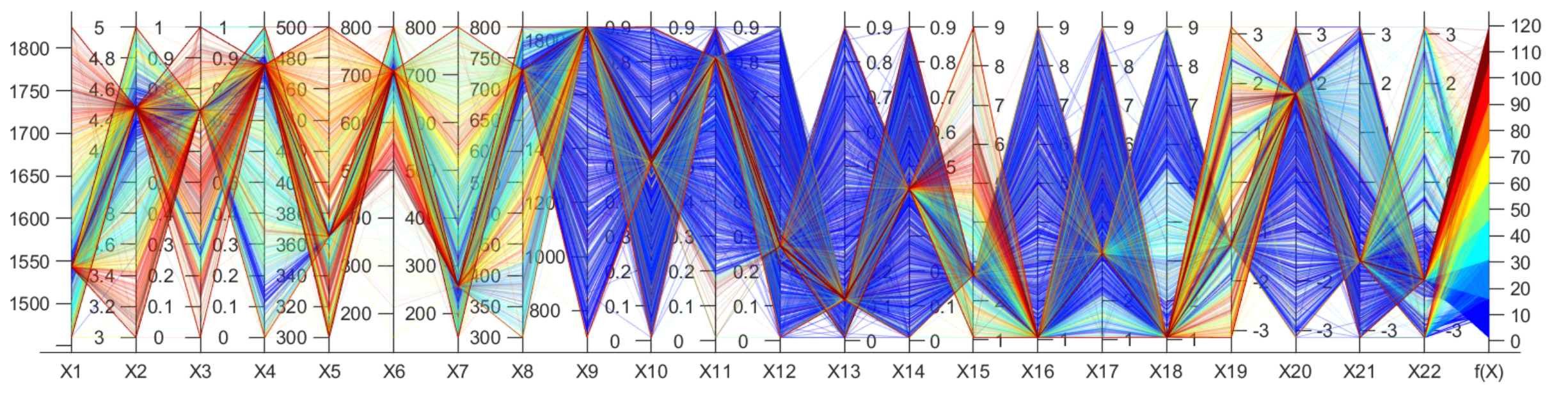}}\\
 \subfloat[]{
 \includegraphics[clip,height=2.7cm,width=1.0\columnwidth]{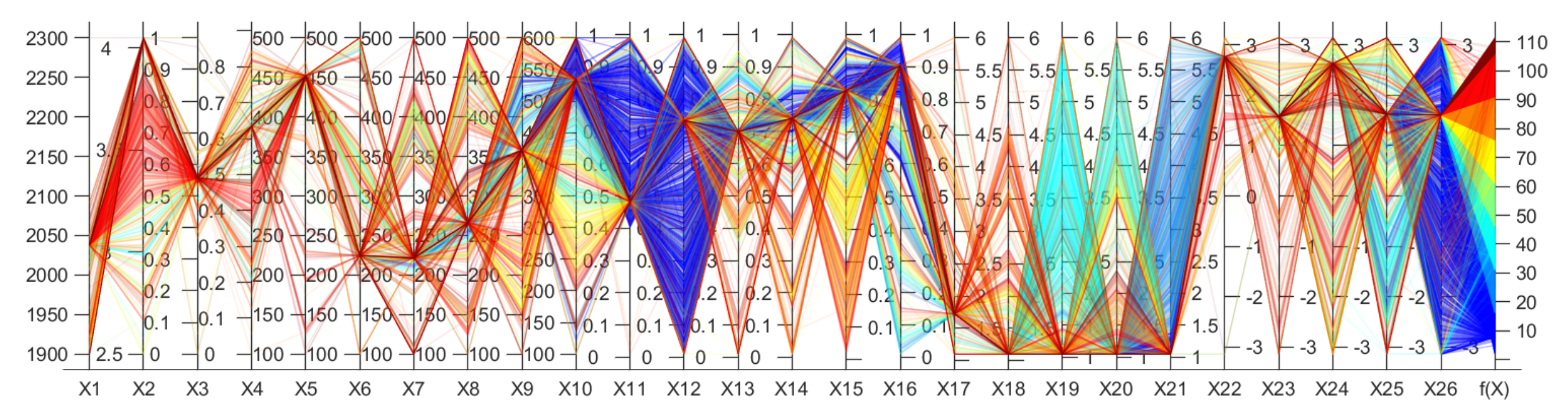}
 }\\
\caption{ Parallel coordinate plots of (a) Cassini1-MINLP, (b) Sagas, (c) Messenger (reduced), (d) Cassini2, (e) Rosetta and (f) Messenger (full).}%
\label{fig:parallel_cooperate_3}%
\end{figure}

In Figures~\ref{fig:parallel_cooperate_1} to~\ref{fig:parallel_cooperate_3} (for the eight single-objective problems), the columns represent the decision variables of the benchmarks, and coloured lines connect the values that belong to a solution vector. The colour denotes the respective solution quality: the most efficient solutions are dark blue. The primary observation of Figure~\ref{fig:parallel_cooperate_2} is that some variables play an important role in optimizing the trajectory cost ($\Delta V$), for instance $X1$ to $X8$ in Cassini2 (Figure~\ref{fig:parallel_cooperate_3}), where small changes result in the decision variable can result in large changes in the objective value. As another example, consider the variable $X3$ in Cassini1 or the variable $X5$ in GTOC1, which are highly sensitive. For practical approaches, this means that they need to be able to focus differently on the different decision variables.

Next, we use a grid search that is a systematic approach with a very small step (mesh) size ($\mu=(Ub-Lb)*0.001$) as a sampling method across the entire search space. The grid search provides a uniform and discrete sampling of the landscape. While systematic, grid searches can be expensive landscape analysis methods where the number of decision variables is high (curse of dimensionality~\cite{bergstra2012random}) or the objective function is computationally expensive. Hence, we limit ourselves to pairs of grid searches, i.e. for each possible pair of decision variables.

\begin{figure}[hbt!]
\centering
\subfloat[]{
 \includegraphics[clip,width=\columnwidth]{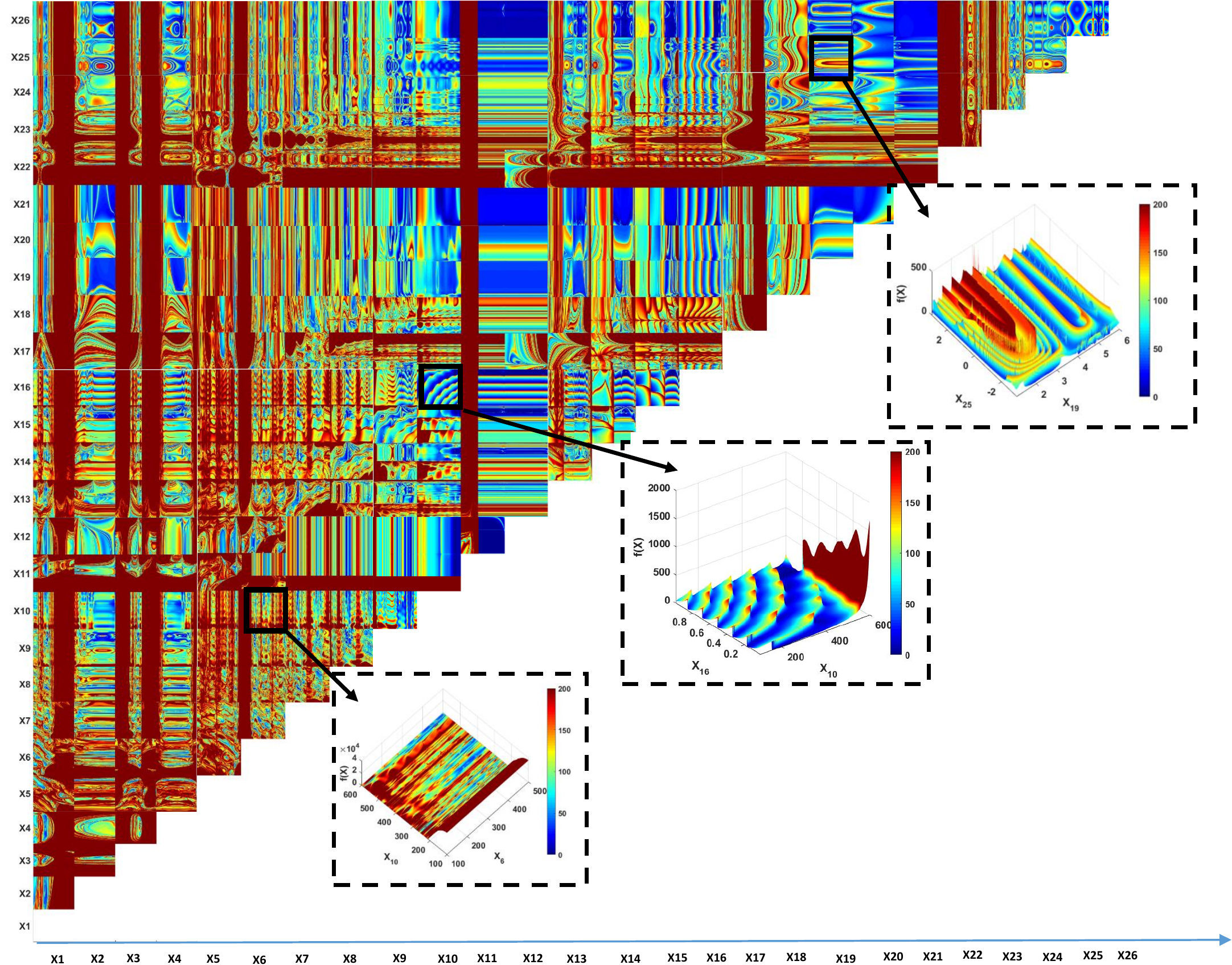}}\\
 \subfloat[]{
 \includegraphics[clip,width=0.49\columnwidth]{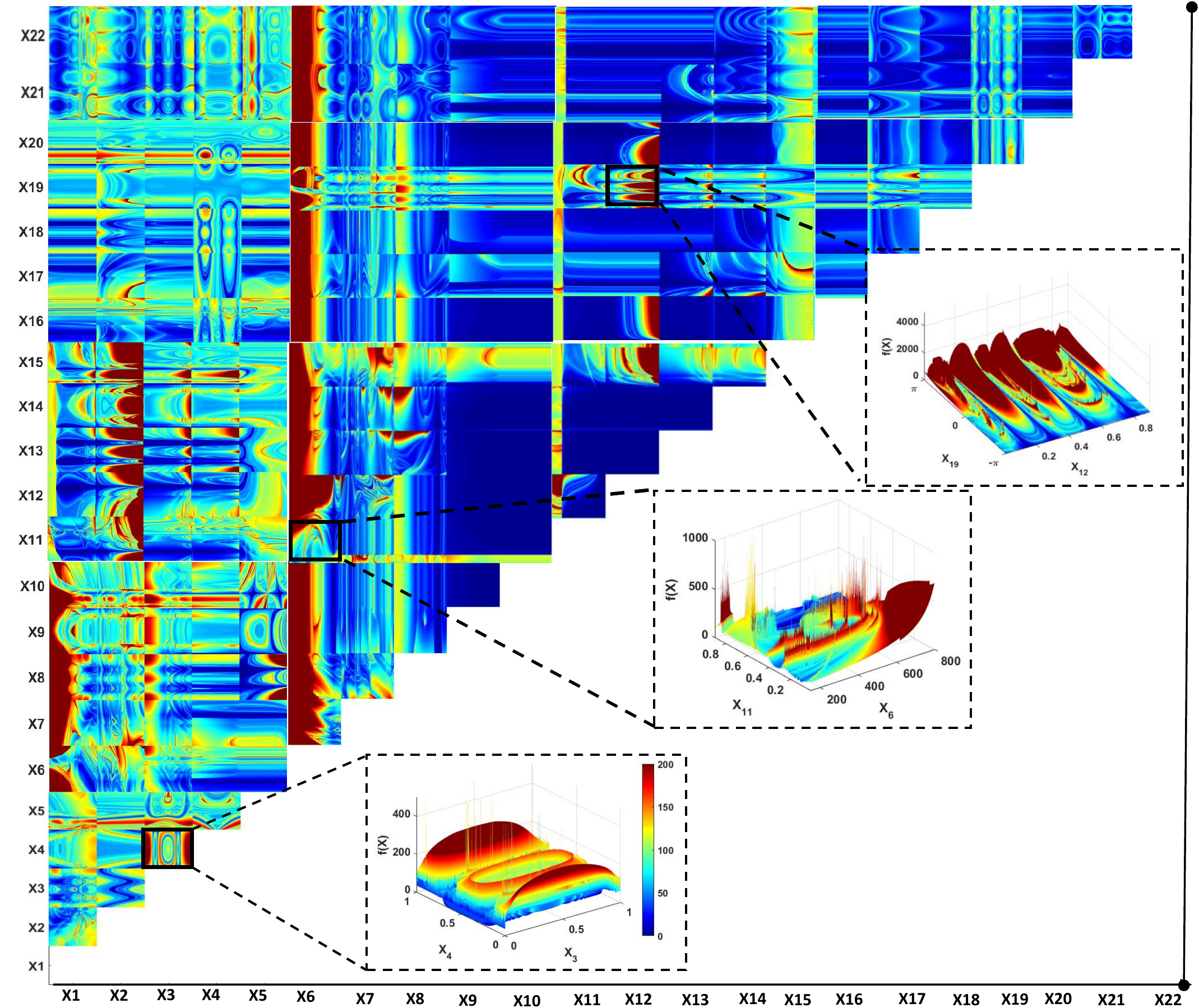}}
 \subfloat[]{
\includegraphics[clip,width=0.5\columnwidth]{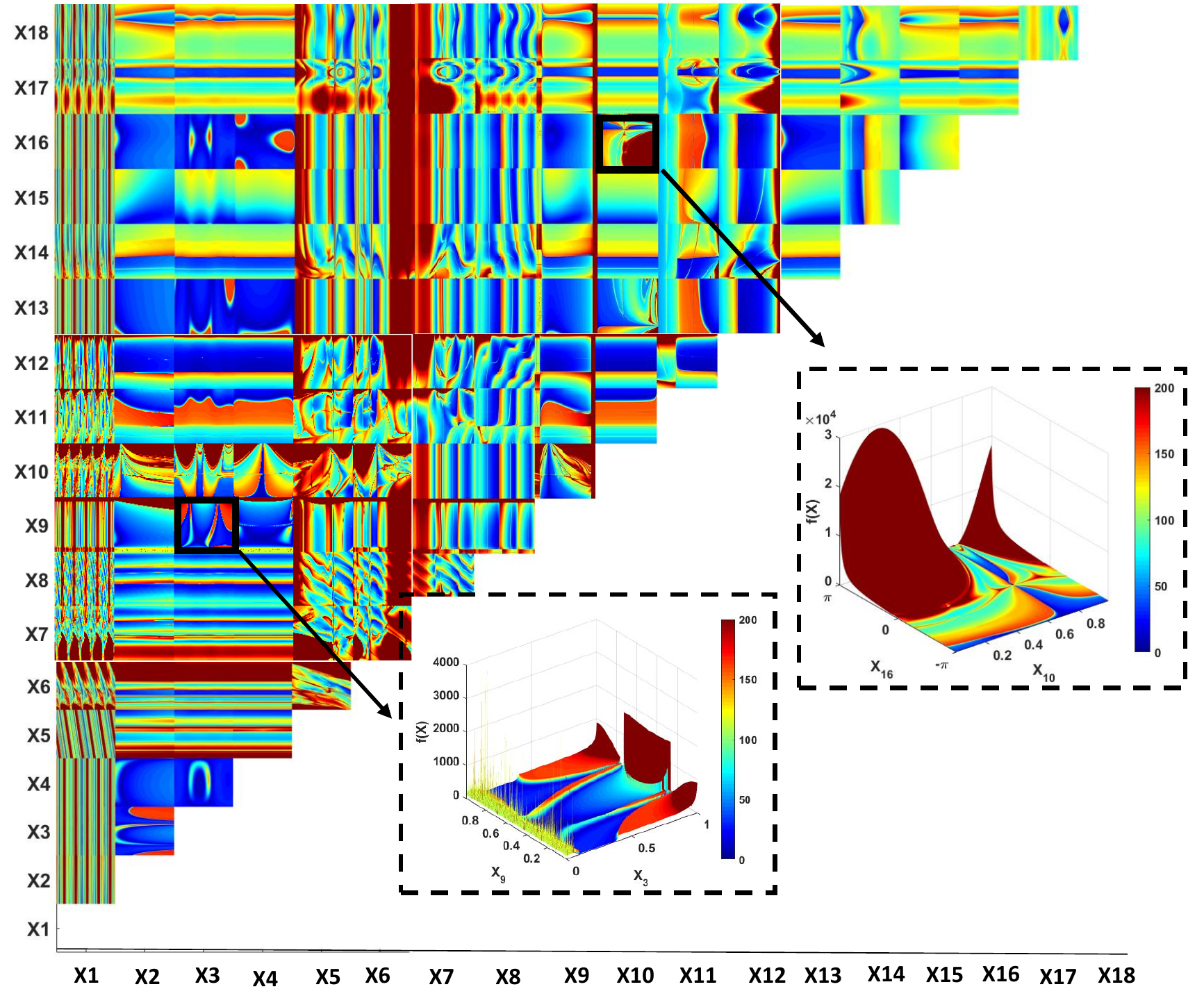}}
\caption{Landscape analysis of GTOPX benchmarks using grid search: (a) Messenger (full), (b) Rosetta, (c) Messenger (reduced).}%
\label{fig:landscape_analysis}%
\end{figure}

Figure~\ref{fig:landscape_analysis} shows results for Messenger (full), Messenger (reduced) and Rosetta. We can see a high level of ruggedness and a large number of spikes in the landscapes. Based on this observed multi-modality (i.e. multiple local optima), it might be worth considering optimization methods in the future such as niching techniques~\cite{casas2015genetic}, crowding~\cite{mengshoel1999probabilistic}, fitness sharing~\cite{goldberg1987genetic}, species conserving~\cite{li2002species} and covariance matrix adaptation~\cite{hansen2001completely}.

\newpage

As a side-effect of our grid search and of the local sampling, we have found small improvements over the previously best known solutions for the Rosetta benchmark. A new solution with a percentage change improvement of 0.00075$\%$ was found. Note that in case of the GTOP benchmarks new solutions must traditionally hold a percentage change improvement of at least 0.1$\%$ to be be considered significant. Nonetheless, other researchers (e.g. \cite{Gregoire}) have published improvements below the $0.1\%$ threshold and thus such new solutions displayed here might be interesting to some.

\begin{table}[ht!]
\caption{New solutions for Rosetta  }
\label{Table:Rosetta}
\small
\centering
    \scalebox{0.8}{
\begin{tabular}{l|l|p{3cm}|p{3cm}}
\hline
\multicolumn{4}{c}{\textbf{\large{Rosetta}}}    \\ \hline \hline
     & Previous Best  & New solution\newline (grid search)& New solution\newline (local search)   \\ \hline
x1   & 1542.802723  & 1542.802723& 1542.802723 \\\hline
x2   & 4.478444171  & 4.478444171& 4.478444171 \\\hline
x3   & 0.73169868   & 0.73169868& 0.73169868   \\\hline
x4   & 0.878289696  & 0.878289696& 0.878289696 \\\hline
x5   & 365.2423131  & 365.2423131& 365.2423131  \\\hline
x6   & 707.7546444  & 707.7546444& 707.7546444 \\\hline
x7   & 257.3238516  & 257.3238516& 257.3238516 \\\hline
x8   & 730.4837236  & 730.4837236& 730.4837236 \\\hline
x9   & 1850         & 1850 &  1850      \\\hline
x10  & 0.469187104  & \textbf{0.51018  }&  \textbf{0.512067}  \\\hline
x11  & 0.810371727  & 0.810371727 & 0.810371727 \\\hline
x12  & 0.057240939  & \textbf{0.25119 }& \textbf{0.2758878 }   \\\hline
x13  & 0.123333369  & 0.123333369& \textbf{0.119192979 }\\\hline
x14  & 0.436535683  & 0.436535683 & 0.43674223\\\hline
x15  & 2.657626174  & 2.657626174 & 2.657626174\\\hline
x16  & 1.05         & 1.05       & 1.05  \\\hline
x17  & 3.197806169  & 3.197806169& 3.197806169 \\\hline
x18  & 1.056221792  & 1.056221792 & 1.056221792\\\hline
x19  & -1.253888118 & -1.253888118& -1.253888118\\\hline
x20  & 1.78760233   & 1.78760233   &1.78760233\\\hline
x21  & -1.594671417 & -1.594671417 &-1.594671417\\\hline
x22  & -1.977325495 & -1.977325495 &-1.977325495\\\hline
f(x) & 1.34335206  & \textbf{1.34334453}&\textbf{1.34334419} \\\hline
\hline
\end{tabular}
}
\end{table}

\clearpage

\section{Software Metadata}

Table \ref{tab:code} displays the metadata information for the GTOPX software.

\begin{table}[H]
\begin{tabular}{|l|p{4cm}|p{8cm}|}
\hline
C1 & Current code version & V 1.0\\
\hline
C2 & Permanent link to code/repository used for this code version & \url{http://www.midaco-solver.com/index.php/about/benchmarks/gtopx} \\
\hline
C3 & Code Ocean compute capsule & na \\
\hline
C4 & Legal Code License   &  GNU General Public License\\
\hline
C5 & Code versioning system used & fixed version upload on webpage \\
\hline
C6 & Software code languages, tools, and services used & Matlab, C/C++ and Python\\
\hline
C7 & Compilation requirements, operating environments \& dependencies & C++ source code, Matlab mex source code, pre-compiled dll/so libraries for Python on Windows, Linux and MacOS\\
\hline
C8 & If available Link to developer documentation/manual & \url{http://www.midaco-solver.com/index.php/about/benchmarks/gtopx}\\
\hline
C9 & Support email for questions &\url{info@midaco-solver.com } \\
\hline
\end{tabular}
\caption{GTOPX software metadata}
\label{tab:code} 
\end{table}

\section{Conclusions}
\label{sec:Conclusions}

With this article, we introduce an extended and refurbished version of ESA's well-known GTOP database. The new version, called GTOPX, includes three new benchmark instances featuring mixed-integer and multi-objective properties. While the original Cassini1 benchmark is clearly the easiest instance, its extensions are more difficult to solve.

A simplified and user-friendly C/C++ source code base with efficient numerical gateways into Python and Matlab make the GTOPX database an attractive choice for researchers wishing to put advanced optimization algorithms to the test. The important feature of thread-safe function calls is provided for all GTOPX benchmarks, allowing the parallel execution of these benchmark functions which is a highly desired feature for modern optimization algorithms. We also provide a first fitness landscape analysis of the single-objective GTOPX benchmarks to visually characterize the relations among decision variables. Given the highly non-linear nature of interplanetary space mission trajectories, which translates into the difficulty to solve these benchmarks, the GTOPX collection provides a challenge for years to come.

\section*{Acknowledgements}

The authors are grateful to the Advanced Concept Team (ACT) of the European Space Agency (ESA) and in particular to Dario Izzo for creating the original GTOP database. We would also like to express our gratitude to the anonymous reviewers who provided valuable comments. This work was supported by JSPS, Japan KAKENHI Grant Number JP20K11967.

\section*{Conflict of Interest}


The authors declare that they have no known competing financial interests or personal relationships that could have appeared to influence the work reported in this paper. 

\end{document}